%% file: iclr2025_conference.tex
\documentclass{article} 
\usepackage{iclr2025_conference,times}

\input{math_commands.tex}

\usepackage{hyperref}
\usepackage{url}
\usepackage{graphicx}
\usepackage{booktabs}
\usepackage{multirow}
\title{Metadata, Wavelet, and Time Aware Diffusion Models for Satellite Image Super Resolution}

\author{Luigi Sigillo, Renato Giamba, Danilo Comminiello \\
Dept. of Information Engineering, Electronics, and Telecomm.,\\
Sapienza University of Rome, Italy\\
\texttt{luigi.sigillo@uniroma1.it} \\
}

\iclrfinalcopy 
\begin{document}

\maketitle
\vspace{-3mm}

\begin{abstract}
\vspace{-3mm}
The acquisition of high-resolution satellite imagery is often constrained by the spatial and temporal limitations of satellite sensors, as well as the high costs associated with frequent observations. These challenges hinder applications such as environmental monitoring, disaster response, and agricultural management, which require fine-grained and high-resolution data. 
In this paper, we propose MWT-Diff, an innovative framework for satellite image super-resolution (SR) that combines latent diffusion models with wavelet transforms to address these challenges.
At the core of the framework is a novel metadata-, wavelet-, and time-aware encoder (MWT-Encoder), which generates embeddings that capture metadata attributes, multi-scale frequency information, and temporal relationships. 
The embedded feature representations steer the hierarchical diffusion dynamics, through which the model progressively reconstructs high-resolution satellite imagery from low-resolution inputs. This process preserves critical spatial characteristics including textural patterns, boundary discontinuities, and high-frequency spectral components essential for detailed remote sensing analysis.
The comparative analysis of MWT-Diff across multiple datasets demonstrated favorable performance compared to recent approaches, as measured by standard perceptual quality metrics including FID and LPIPS. The code is available at \url{https://github.com/LuigiSigillo/MWT-Diff}
\end{abstract}

\section{Introduction}
\textbf{Satellite Imaging.} 
The need for high-resolution (HR) satellite imagery has grown exponentially with applications spanning environmental monitoring, urban planning, disaster response, and agricultural management \citep{10371219}. However, the acquisition of satellite images, particularly high-resolution imagery, can be both costly and limited by the spatial and temporal resolution constraints of satellite sensors \citep{khanna2024diffusionsat}. For instance, Sentinel-2 satellites provide images of Earth's land surface every five days, with resolutions ranging from 10 to 60 meters \citep{cornebise2022open}. While valuable, this level of detail often falls short for tasks requiring fine-grained monitoring, such as assessing agricultural health or managing wildfires. Acquiring high-resolution images, such as those captured by SPOT6, for large areas and at high frequencies remains a significant challenge due to cost and technical limitations.

\textbf{Multi-Image and Hyperspectral Image Super-Resolution.} Image super-resolution (SR) has emerged as a pivotal area in computer vision, aiming to enhance the resolution of low-quality images to generate high-resolution outputs \citep{wang2020deep}, this has been also exploited in LR satellite images \citep{cornebise2022open}. \citet{khanna2024diffusionsat} introduced DiffusionSat which represents an important step towards this task which allows the use of LR image sequences as input, along with metadata, to generate high-quality HR output. These inputs often consist of multispectral satellite images captured at varying ground sampling distances (GSDs) e.g. it can utilize LR images from Sentinel-2 to produce HR images comparable in quality to those from HR datasets like fMoW \citep{christie2018functional}.
Another milestone in the topic is SatDiffMoe \cite{luo2024satdiffmoe} which leverages the complementary information embedded in multiple LR satellite images captured at different time points. The key innovation lies in its diffusion-based fusion algorithm, which conditions the reconstruction of HR images on each individual LR input while merging their outputs to achieve enhanced detail and accuracy.

\textbf{SR exploiting prior knowledge.} To enhance the quality of generated images, many SR techniques have incorporated prior knowledge to guide the reconstruction process. Reference-based SR methods leverage similar HR images as explicit priors to assist in generating the corresponding HR outputs \citep{zheng2018crossnet}. However, aligning the features of reference images with the LR input is challenging in real-world scenarios, and such explicit priors are not always readily available, limiting their practicality. Recent SR approaches have shifted to implicit priors, which offer more flexibility and generalization \citep{wang2018recovering, yu2018crafting, zhao, Menon2020PULSESP}. 
However, these methods often suffer from insufficient representational capacity or inaccurate feature alignment, resulting in suboptimal output quality.
StableSR \citet{wang2024exploiting} represents a significant advancement by leveraging the robust and extensive generative prior found in pre-trained diffusion models, fine-tuning a frozen pre-trained diffusion model with only a small number of trainable parameters, making it both efficient and highly effective. 


In this paper we explore latent diffusion models within the problem of satellite image SR, establishing a robust framework capable of reconstructing HR images from LR input while addressing the unique characteristics and challenges of satellite imagery. The proposed methodology integrates wavelet transforms and conditional generation, to better capture fine details and contextual relationships within the data.

\section{The Proposed MWT-Diff Method}
\label{sec:method}
\textbf{Feature Modulation.} The conditional generation approach for SR with diffusion models leverages the latent representation of degraded input images, extracted through an encoder, to guide the generation process.
To enhance the guidance during generation, in StableSR \citet{wang2024exploiting} an additional encoder is introduced to extract multi-scale features \(\{F^n\}_{n=1}^N\) from the degraded LR image. These extracted features were subsequently used to modulate the intermediate feature maps, \(\{F_{\text{dif}}^n\}_{n=1}^N\), within the residual blocks of Stable Diffusion (SD) \citep{Rombach2021HighResolutionIS} via spatial feature transformations (SFT) introduced by \citet{wang2018recovering}.



During fine-tuning, the weights of the SD model are frozen, and only the encoder and the SFT layers are trained. This approach ensured the preservation of the generative prior inherent in SD while allowing structural information from the LR image to be incorporated effectively.  

Additionally, the capacity of the model to integrate textual information is improved by utilizing CLIP-encoded text embeddings during training. This enhancement facilitates a more nuanced understanding of both the image category and the country metadata embedded in the dataset.

\textbf{MWT-Encoder.} We introduce a metadata- and wavelet- and time-aware encoder (MWT-Encoder) as the core of the baseline model architecture inspired by \citet{shipin} that introduced a class- and time-aware encoder, and \citet{khanna2024diffusionsat} which highlighted the importance of metadata (e.g., timestamp, latitude, longitude) associated with satellite images.   

In this context, each image \(x \in \mathbb{R}^{3 \times H \times W}\) is paired with numerical metadata \(K \in \mathbb{R}^M\), where \(M\) represents the number of metadata items. Following the methodology introduced in StableSR, we incorporate temporal information into the encoder using a time-embedding layer. 
We encode metadata using sinusoidal timestep embeddings, as in \citet{khanna2024diffusionsat}, to avoid discretization errors from CLIP’s poor numerical encoding \citep{radford2021learning}.

\begin{figure*}[!t]
    \centering
    \includegraphics[width=0.85\linewidth]{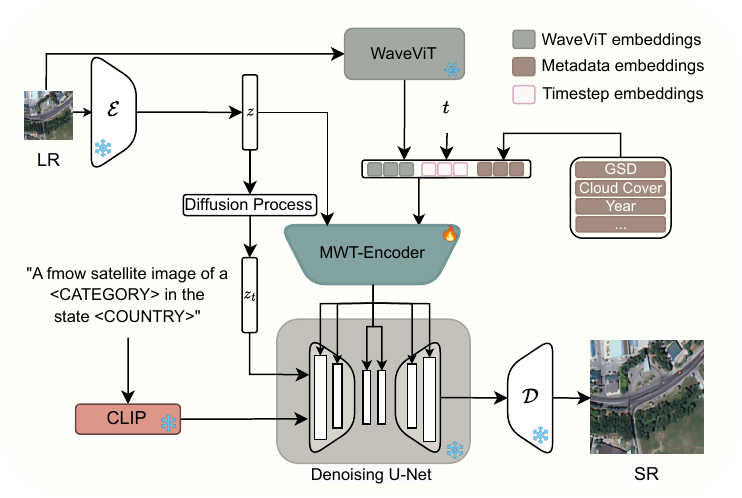}
    \caption{Overview of MWT-Diff framework with our MWT-Encoder.}
    \label{fig:architecture}
\end{figure*}

Recent works show that the use of wavelets can improve SR within diffusion models \citep{aloisi2024waveletdiffusionganimage, zhao2024wavelet}. Our idea is to take advantage of the wavelets, not to perform diffusion in frequency space \citep{phung2023wavelet}, but instead to extract features from the wavelet transform of the satellite images through the pre-training of a Wavelet Vision Transformer (WaveViT) \citep{wavevit2022}.

WaveViT consists of four stages, each comprising a patch embedding layer, a series of Wavelet blocks, and subsequent feed-forward layers. In the initial phase of the architecture, discrete wavelet transforms (DWT) are applied to the input images. This operation decomposes the images into multiple frequency components, enabling the model to simultaneously capture both low-frequency (global structure) and high-frequency (fine detail) information. This is particularly critical for preserving the intricate textures and details commonly found in satellite imagery.
After pre-training, which extracts robust feature representation leveraged throughout the network to perform the super-resolution task, we integrate WaveViT in our architecture, concatenating the embeddings along with the timesteps and metadata embeddings. 



The concatenated metadata tensors \(m \in \mathbb{R}^{1024}\), wavelet embeddings \(w \in \mathbb{R}^{1024}\), and timestep embeddings \(t \in \mathbb{R}^{1024}\) yield a vector \(b \in \mathbb{R}^{3072}\), which is combined with the latent representation of the image \(z \in \mathbb{R}^{h \times w \times c}\). The latent \(z\) is obtained from the encoder \(z = \mathcal{E}(x)\), where \(x \in \mathbb{R}^{3 \times H \times W}\) is the low-resolution image. Both the encoder \(\mathcal{E}\) and the decoder \(\mathcal{D}\) are frozen components of the VAE used in SD \cite{Rombach2021HighResolutionIS}, alongside the time-conditional U-Net responsible for performing the latent diffusion process.  

The MWT-Encoder \(\delta_\theta\) generates conditioning embeddings that guide the denoising process of the conditional auto-encoder \(\epsilon_\theta\) at multiple scales. These embeddings are incorporated following the architecture of the U-Net.
The complete objective function for training the model is defined as:  

\begin{equation}
\label{eq:training_obj}
     L = \mathbb{E}_{E(x),y, \epsilon \sim \mathcal{N}(0,1),m,w,t} \left[ \left\| \epsilon - \epsilon_\theta(z_t, \delta_\theta(m, w, t, z), \tau(y)) \right\|_2^2 \right]
\end{equation}

In equation \ref{eq:training_obj}, \(\tau(y)\) represents the domain-specific encoder, implemented using a frozen CLIP model, which processes the conditioning caption \(y\). The variable \(m\) corresponds to the metadata embeddings, \(w\) is the wavelet embedding, \(t\) is the timestep embedding, \(z\) is the latent representation of the low-resolution image extracted by the VAE encoder, and \(z_t\) is the latent representation after the addition of noise during the diffusion process.  

The objective is to optimize the MWT-Encoder \(\delta_\theta(m, w, t, z)\) to provide effective conditioning embeddings. These embeddings ensure that the denoising process generates HR images that faithfully capture the information encoded in both the input metadata and the conditioning captions.
The final MWT-Diff model used for the super-resolution task is illustrated in Figure \ref{fig:architecture} within our MWT-Encoder.
\section{Experimental Results}
\label{sec:experiment}
\begin{figure*}[h!]
    \centering
    \includegraphics[width=\linewidth]{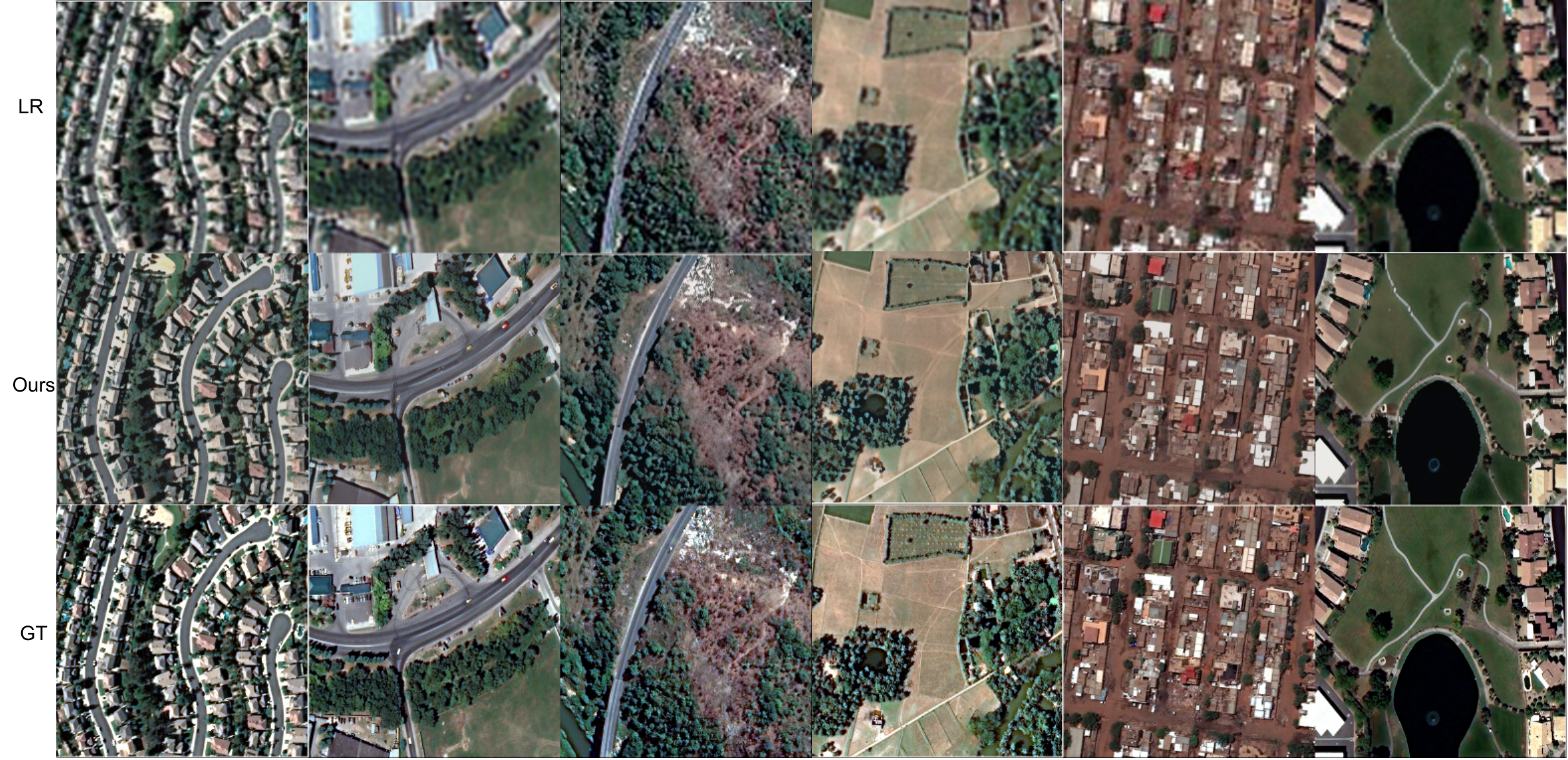}
    \caption{Qualitative results on 128x128 $\rightarrow$ 512x512 with fMoW \citep{christie2018functional}.}
    \label{fig:fmow_results}
\end{figure*}


We evaluate our model on two datasets. The first dataset is a subset of the fMoW dataset \citet{christie2018functional}, used for the super-resolution task of 4x scaling, where 128x128 pixel low-resolution images are upscaled to 512x512 pixel high-resolution images. The second dataset is a subset of the Sentinel2-fMoW dataset \citet{cong2022satmae}, employed for the super-resolution task on hyperspectral images. For inference we adopt the DDPM \cite{ho2020denoising} sampling strategy utilizing $200$ timesteps.


\begin{table}[t] 
\centering 

\label{tab:bench_acc}
\begin{tabular}{llccc} 
\toprule
    & Model & FID $\downarrow$ & LPIPS $\downarrow$ \\
\midrule
\multirow{3}{*}{\rotatebox{90}{\textbf{fMoW}}}
& Low Resolution  & 114.38 & 0.756 +/- 0.004 \\
   & StableSR  & 53.85 & 0.345 +/- 0.002\\
   & \textbf{MWT-Diff} & \textbf{53.07} \textcolor{blue}{(-1.44\%)} & \textbf{0.336} +/- 0.002 \textcolor{blue}{(-2.61\%)} \\

\midrule
\multirow{8}{*}{\rotatebox{90}{\textbf{Sentinel2-fMoW}}} 
    & WorldStrat \cite{cornebise2022open}  & 426.7 & 0.736 $\pm$ 0.092 \\
       & MSRResNet \cite{wang2018esrgan}  & 286.5 & 0.783 $\pm$ 0.081 \\
       & DBPN \cite{haris2018deep} & 278.2 & 0.750 $\pm$ 0.052 \\
   & Pix2Pix \cite{isola2017image} & 196.3 & 0.643 $\pm$ 0.045 \\
  &  SatDiffMoE \cite{luo2024satdiffmoe} & 115.6 & 0.606 $\pm$ 0.044 \\
  &  DiffusionSat \cite{khanna2024diffusionsat} & 102.9 & 0.638 $\pm$ 0.034 \\
     & ControlNet \cite{zhang2023adding} & 102.3 & 0.644 $\pm$ 0.034 \\
   & \textbf{MWT-Diff} & \textbf{98.1} \textcolor{blue}{(-4.11\%)} & \textbf{0.555} $\pm$ 0.003 \textcolor{blue}{(-8.41\%)} \\
\bottomrule
\end{tabular}
\vspace{-1mm}
\caption{Evaluation metrics comparison for the $128\times128 \rightarrow 512\times512$ super resolution task on fMoW dataset \cite{christie2018functional}, and on Sentinel2-fMoW dataset \cite{cong2022satmae}.
}
\end{table}

The results underscore the superior performance of our model across both the fMoW and hyperspectral image super-resolution tasks, as presented in Table \ref{tab:bench_acc}. By integrating our MWT-Encoder, the model significantly enhances its ability to reconstruct intricate features and capture fine-grained details within the images. 
MWT-Diff consistently outperforms its counterparts, including traditional CNN-based methods and other diffusion models, as evidenced by its superior performance in both FID and LPIPS metrics.


\section{Conclusion}
\label{sec:conclusion}
In conclusion, this paper presents a novel latent diffusion model to address the challenge of satellite image super-resolution. By incorporating the MWT-Encoder, the proposed framework effectively utilizes contextual information, leading to improved image fidelity and detail.
Our MWT-Diff achieved superior performance with lower FID and LPIPS scores, demonstrating its ability to generate high-quality, ground-truth-like images for practical applications in satellite imaging, including urban planning, environmental monitoring, and disaster management.
\section*{Acknowledgments}
This work was partly supported by ``Ricerca e innovazione nel Lazio - incentivi per i dottorati di innovazione per le imprese e per la PA - L.R. 13/2008” of Regione Lazio, Project ``Deep Learning Generativo nel Dominio Ipercomplesso per Applicazioni di Intelligenza Artificiale ad Alta Efficienza Energetica”, under grant number 21027NP000000136, and by the European Union under the Italian National Recovery and Resilience Plan (NRRP) of NextGenerationEU, ``Rome Technopole” (CUP B83C22002820006)—Flagship Project 5: ``Digital Transition through AESA radar technology, quantum cryptography and quantum communications”.
\bibliography{iclr2025_conference}
\bibliographystyle{iclr2025_conference}
\newpage
\appendix
\section{Appendix}

\begin{figure*}[h!]
    \centering
    \includegraphics[width=\linewidth]{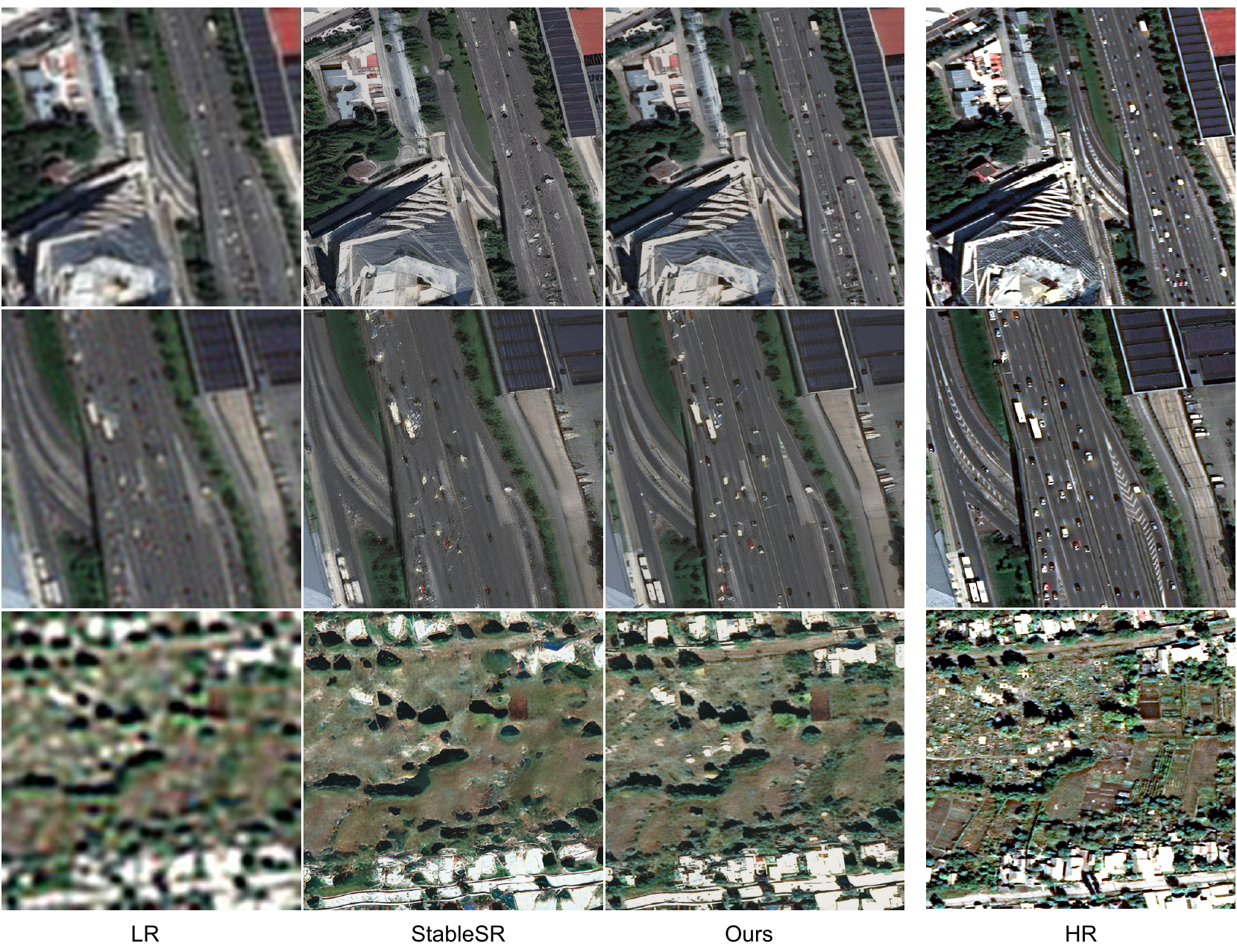}
    \caption{Additional results on 128x128 $\rightarrow$ 512x512 with fMoW \cite{christie2018functional}. Comparison between MWT-Diff and StableSR.}
    \label{fig:fmow_results_2}
\end{figure*}

\subsection{Dataset details}
\subsubsection{Functional Map of the World (fMoW)}
The metadata accompanying each image of the fMoW dataset \cite{christie2018functional} provides contextual information, including location, time, sun angles, physical sizes, and other attributes, facilitating a more informed analysis of the objects within the images.

The original fMoW dataset consists of over 1 million satellite images collected from more than 200 countries, each annotated with at least one bounding box corresponding to one of 63 categories.
For the purpose of conducting a 4x scaling super-resolution task—transforming 128×128 pixel LR images into 512×512 pixel HR images we select a balanced subset of 100k samples from the original dataset, ensuring proportional representation of all 63 classes.

To generate the HR images, a resize operation to 512×512 pixels followed by a center crop was applied, preserving the critical details of the images. The LR images were derived from the HR counterparts by degrading them using various techniques outlined Real-ESRGAN \cite{wang2021real}, effectively reducing their resolution while simulating realistic degradations.

Additionally, to establish a baseline for comparison with the super-resolution model outputs, a standard upscaling technique was applied to the LR images, creating reference images for evaluation. The resulting refined dataset is optimized and prepared for training and benchmarking super-resolution models.
\subsubsection{Functional Map of the World - Sentinel-2 (fMoW-Sentinel)}
The dataset \cite{cong2022satmae} comprises temporal image series collected by the Sentinel-2 satellite, corresponding to the geographic locations of the original fMoW dataset. These images were captured at multiple time points and are categorized into 62 distinct types of building and land use, following the classification scheme of the fMoW dataset.  

The fMoW-Sentinel dataset consists of Sentinel-2 images with a 10-meter spatial resolution, created from cloud composites aggregated over 90-day intervals. Each image includes 13 spectral bands from the Sentinel-2 surface reflectance dataset, offering a comprehensive multispectral representation.
Building on previous works \cite{cong2022satmae, luo2024satdiffmoe}, we use a derived dataset named fMoW-Sentinel-fMoW-RGB. This dataset pairs Sentinel-2 images (10m–60m ground sampling distance, GSD) with high-resolution fMoW-RGB images (0.3m–1.5m GSD) at each of the original fMoW locations.  

A key distinction from SatdiffMoE is the handling of high-resolution images. While it uses bounding boxes provided in the metadata to extract relevant areas and crop them into 512×512 patches, the fMoW high-resolution images in this work were resized in their entirety to 512×512 pixels.  

For each high-resolution fMoW image, we select one corresponding Sentinel-2 image from the same location. Similar to SattdiffMoE, only the RGB bands of the Sentinel-2 images were retained. These RGB images were then resized to match the dimensions of the fMoW-RGB images, ensuring consistency in resolution for downstream tasks.  

\subsection{Training details}
The MWT-Diff model is trained for eight and four epochs, respectively. Further training beyond these limits resulted in worsening FID scores. The Adam optimizer was employed, with a learning rate set to 5e-5. An instance of the WaveViT-B model was pre-trained and evaluated directly on the fMoW dataset. This evaluation was conducted on 1,000 images from the fMoW validation set.

\subsection{Evaluation details}
In line with the methodology presented by \cite{luo2024satdiffmoe}, the evaluation of the WaveViT is further refined by focusing on the first $100$ images from the validation set across six specific classes: Airport, Amusement Park, Car Dealership, Crop Field, Educational Institution, and Electric Substation.
We generate a synthetic set of captions for each class of the fMoW dataset using the pattern: "A fMoW satellite image of a class\_type". Each image is paired with these synthetic captions and input into the model under evaluation. The model then predicted the most relevant caption associated with each image.

\subsection{Additional Qualitative Comparisons}
\begin{figure*}[h!]
    \centering
    \includegraphics[width=\linewidth]{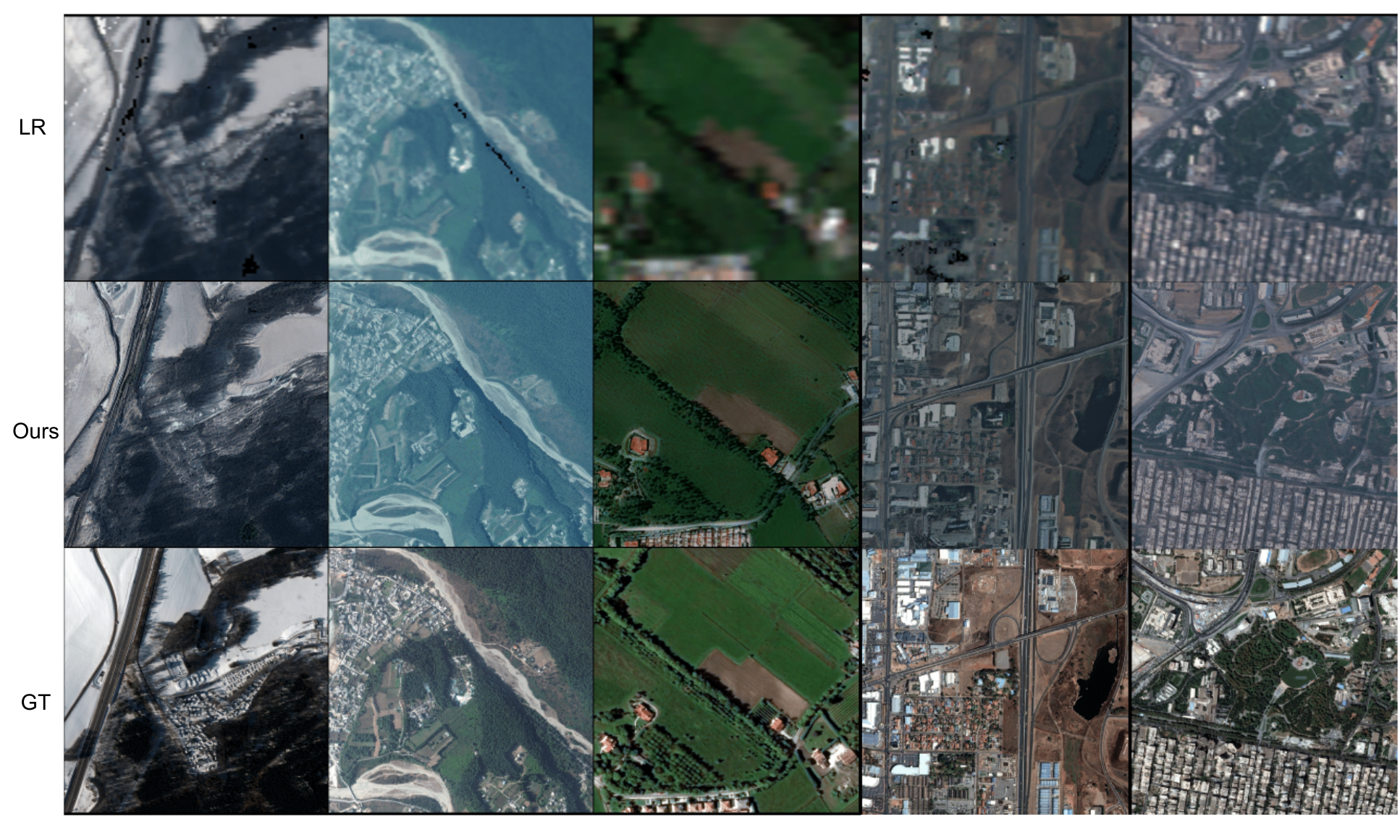}
    \caption{Qualitative results on Sentinel2-fMoW dataset \cite{cong2022satmae}. The image includes a comparison of the LR Sentinel-2 input, the output of the model, and the corresponding HR fMoW images.
    }
    \label{fig:sentinel_results}
\end{figure*}

Figure~\ref{fig:sentinel_results} illustrates qualitative results on the Sentinel2-fMoW dataset \cite{cong2022satmae}, showcasing the low-resolution Sentinel-2 input, the super-resolved output generated by MWT-Diff, and the corresponding high-resolution fMoW reference image. Figure~\ref{fig:fmow_results_2} depicts an additional comparison on the standard fMoW dataset \cite{christie2018functional}, where MWT-Diff is directly compared against the StableSR baseline.

The results demonstrate consistent qualitative improvements across datasets. In the Sentinel2-fMoW case, MWT-Diff successfully reconstructs fine-grained structures despite the challenging resolution gap between Sentinel-2 inputs and fMoW ground truth. Meanwhile, in the fMoW comparison, MWT-Diff outputs exhibit clearer boundaries and more coherent textures than those produced by StableSR, indicating the effectiveness of our metadata- and frequency-aware conditioning mechanisms.

These examples highlight the model’s generalization capability and the robustness of its generative priors across different spatial resolutions and sensor characteristics.

\subsection{Ablation Study}
\label{sec:ablation}

We conduct a series of ablation experiments to evaluate the individual contributions and trade-offs of key components in the MWT-Diff framework. This analysis explores the impact of removing conditioning modules and examines design choices that affect the model's complexity, efficiency, and applicability.

\begin{figure}[h!]
\centering
\includegraphics[width=0.9\textwidth]{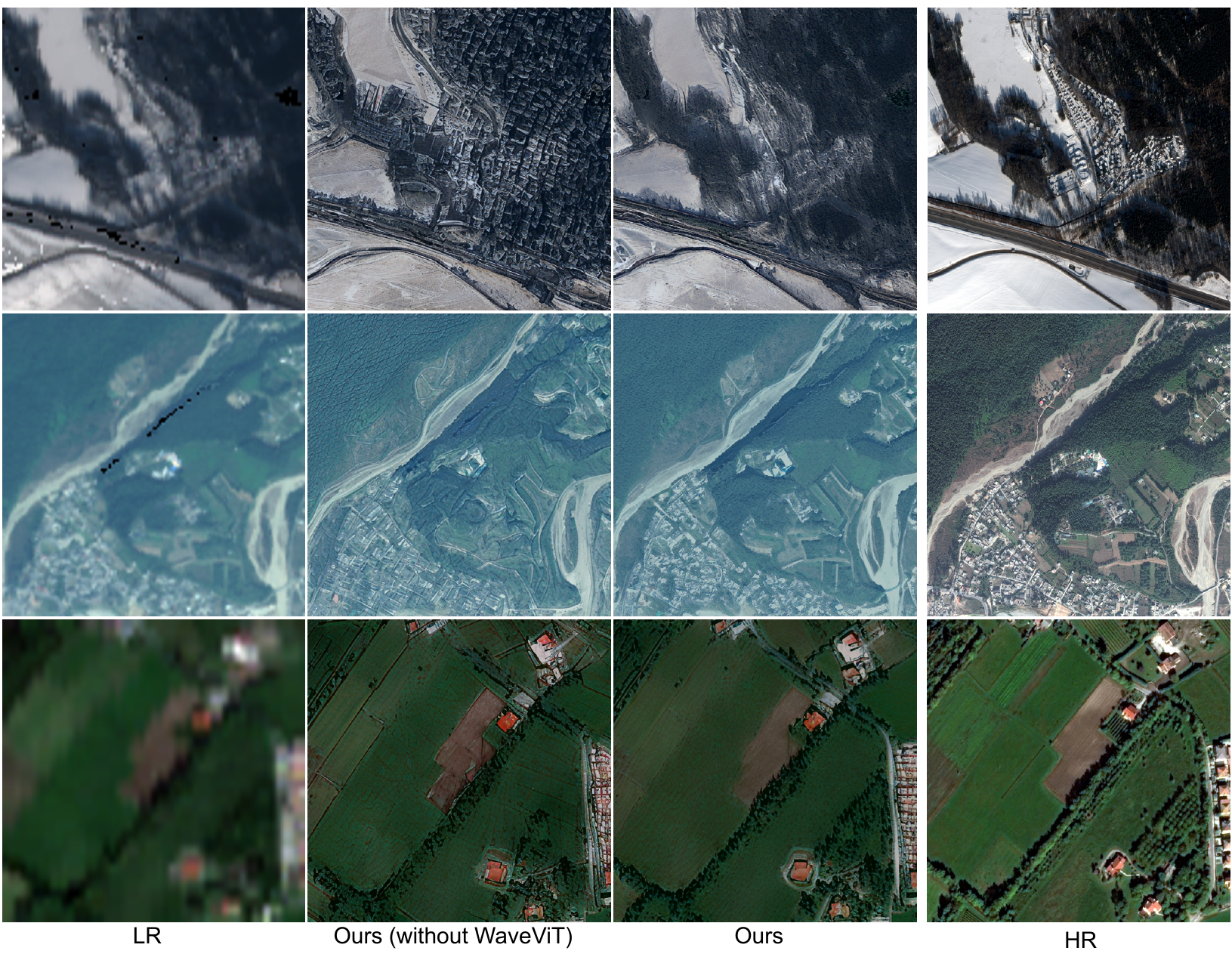} 
\caption{Qualitative comparison of super-resolution outputs with and without WaveViT embeddings. From left to right: Low-resolution input, MWT-Diff without WaveViT, MWT-Diff with WaveViT, and Ground Truth.}
\label{fig:wavevit_ablation}
\end{figure}
\subsubsection{CLIP-based Text Encoder}

To assess the necessity of the CLIP-based caption encoder, we removed this component from the architecture and retrained the model. The goal was to investigate whether this simplification would reduce computational overhead or accelerate inference. Empirical results indicated no measurable improvement in sampling speed. On the contrary, a slight degradation in performance was observed across FID and LPIPS metrics. These findings suggest that while the CLIP encoder introduces additional complexity, it provides marginal gains in perceptual quality and semantic alignment, which justify its inclusion in the final model.

\subsubsection{Comparison with StableSR and Implications for Real-World Deployment}

We further examined the computational cost of MWT-Diff relative to StableSR. Despite the integration of additional embeddings (metadata, wavelet, and temporal encodings), the inference time remained comparable. This parity is attributed to the lightweight nature of the MWT-Encoder and the use of a single-image input pipeline. Notably, when compared to methods that requires multi-image fusion \citep{khanna2024diffusionsat, luo2024satdiffmoe, cornebise2022open} from different timestamps, MWT-Diff demonstrates superior practicality. The absence of complex temporal alignment and fusion procedures enables more straightforward deployment in real-world scenarios where temporally-aligned sequences may not be available.

\subsubsection{Impact of RGB-Only Input}

Our current implementation utilizes only the RGB bands of the Sentinel-2 imagery. This choice was made to maintain a lean architecture, facilitating efficient training and inference while enhancing the usability of the model in operational environments with limited multispectral data. Although RGB data suffice for a broad range of super-resolution tasks, we recognize the potential advantages of incorporating additional spectral channels. Future work will investigate the integration of multispectral and hyperspectral inputs to assess their impact on reconstruction accuracy and downstream utility, particularly in applications such as vegetation monitoring and land use classification.

\subsubsection{Effect of Wavelet-Based Embeddings (WaveViT)}

To evaluate the contribution of frequency-aware representations, we conducted an experiment disabling the WaveViT component. Figure~\ref{fig:wavevit_ablation} shows a qualitative comparison of super-resolved samples with and without WaveViT-based embeddings.

From visual inspection, it is evident that the inclusion of WaveViT significantly enhances the perceptual sharpness and texture fidelity of the generated images. Samples generated with WaveViT appear less blurry and less noisy, exhibiting structural characteristics that more closely resemble the ground truth. These results confirm that multi-scale frequency features, as captured by WaveViT, play a critical role in recovering fine details and reducing generative artifacts.

\end{document}

%% file: math_commands.tex
\usepackage{amsmath,amsfonts,bm}

\def\eqref#1{equation~\ref{#1}}

\def\1{\bm{1}}

\DeclareMathAlphabet{\mathsfit}{\encodingdefault}{\sfdefault}{m}{sl}
\SetMathAlphabet{\mathsfit}{bold}{\encodingdefault}{\sfdefault}{bx}{n}

